\documentclass{article}

\usepackage{microtype}
\usepackage{graphicx}
\usepackage{subcaption}
\usepackage{booktabs}

\usepackage{hyperref}

\usepackage[preprint]{icml2026}

\usepackage{amsmath}
\usepackage{amssymb}
\usepackage{amsfonts}
\usepackage{mathtools}
\usepackage{amsthm}
\usepackage{nicefrac}
\usepackage{xcolor}
\usepackage{multirow}
\usepackage{makecell}
\usepackage{algorithm}
\usepackage{algorithmic}
\usepackage{url}

\usepackage[capitalize,noabbrev]{cleveref}

\newcommand{\Ldiff}{\mathcal{L}_{\text{diff}}}
\newcommand{\Ljepa}{\mathcal{L}_{\text{JEPA}}}
\newcommand{\Ltotal}{\mathcal{L}_{\text{total}}}
\newcommand{\ztl}{z_{t_L}}
\newcommand{\zth}{z_{t_H}}
\newcommand{\xtl}{x_{t_L}}
\newcommand{\xth}{x_{t_H}}

\icmltitlerunning{DLLM-JEPA: JEPA for Masked Diffusion Language Models}

\begin{document}

\twocolumn[
\icmltitle{DLLM-JEPA: Joint Embedding Predictive Architectures for\\
Masked Diffusion Language Models}

\icmlsetsymbol{equal}{*}

\begin{icmlauthorlist}
\icmlauthor{Sangdae Nam}{vessl}
\end{icmlauthorlist}

\icmlaffiliation{vessl}{VESSL AI}

\icmlcorrespondingauthor{Sangdae Nam}{lucas@vessl.ai}

\icmlkeywords{Joint Embedding Predictive Architectures, Masked Diffusion Language Models, Representation Learning, Fine-Tuning}

\vskip 0.3in
]

\printAffiliationsAndNotice{}

\begin{abstract}
We introduce \textbf{DLLM-JEPA}, a JEPA formulation for masked diffusion language models. JEPA objectives have so far been applied to autoregressive LMs at the cost of explicit paired views and multiple gradient passes per training step. By leveraging the diffusion noise schedule, DLLM-JEPA constructs two views from a single input without requiring paired data, and reduces JEPA training FLOPs by \textbf{33\%} relative to LLM-JEPA's two-gradient-view design through a single gradient pass per step.

Across four tasks and two diffusion backbones, DLLM-JEPA consistently improves over diffusion-only fine-tuning, with modest gains in stable settings (e.g., \textbf{+1.8\,pp} on GSM8K) and larger improvements under more aggressive fine-tuning, while also tightening seed-to-seed variance on the high-variance LLaDA-8B GSM8K cells. In addition, it does not degrade base-model performance on a held-out diffusion-loss probe nor on a small MMLU sanity check.

We further analyze the representation dynamics induced by the objective and observe a consistent empirical pattern: models trained with DLLM-JEPA exhibit larger geometric drift from their pre-trained initialization while maintaining comparable or lower functional forgetting.

These results suggest that DLLM-JEPA provides an efficient way to incorporate representation-level objectives into diffusion language model fine-tuning.
\end{abstract}

\section{Introduction}

The dominant paradigm for training large language models relies on input-space reconstruction: autoregressive (AR) next-token prediction~\citep{brown2020language,touvron2023llama} or masked token reconstruction~\citep{devlin2019bert}.
In contrast, vision has increasingly adopted Joint Embedding Predictive Architectures (JEPAs)~\citep{assran2023ijepa,bardes2024vjepa}, which learn representations by predicting embeddings of one view from another---operating entirely in latent space.
JEPAs have been shown to learn richer, more abstract representations by avoiding pixel-level reconstruction biases~\citep{lecun2022path,littwin2024jepa}.

The recent LLM-JEPA~\citep{huang2025llmjepa} represents a first step toward bringing JEPA objectives to language models.
By treating (text, code) pairs as two views of the same underlying knowledge, LLM-JEPA adds a JEPA loss alongside the standard next-token prediction objective.
Despite promising results, LLM-JEPA faces two fundamental limitations rooted in the autoregressive architecture:

\begin{enumerate}
    \item \textbf{Explicit view requirement.} LLM-JEPA requires datasets with natural two-view structures (e.g., natural language paired with code). The authors themselves note this as a key limitation: \emph{``developing a mechanism akin to data-augmentation in vision would enable JEPA objectives to be used on any dataset''}~\citep{huang2025llmjepa}.

    \item \textbf{Computational overhead from unidirectionality.} AR models process tokens left-to-right with causal masking. To obtain independent embeddings for two views, LLM-JEPA requires a custom block-causal attention mask and two forward passes---both carrying gradients---resulting in $\sim$2$\times$ the training compute of standard fine-tuning.
\end{enumerate}

We observe that \emph{masked diffusion language models}~\citep{nie2025llada,sahoo2024simple,shi2024simplified} resolve both limitations naturally.
Models such as LLaDA~\citep{nie2025llada} employ bidirectional attention and learn to denoise randomly masked tokens---a process that is structurally analogous to JEPA's view prediction.
This insight leads to our proposed method, \textbf{DLLM-JEPA}, whose contributions can be summarized as (1) an \emph{architecturally efficient} adaptation of JEPA to masked-diffusion LMs, (2) a \emph{consistent empirical pattern} in how JEPA reshapes representations in bidirectional LMs, and (3) consistent task gains together with maintained base-capability:

\begin{itemize}
    \item \textbf{A JEPA formulation that saves 33\% of the training FLOPs.} The bidirectional attention of masked-diffusion models lets us generate both JEPA views from the same input via different masking rates (no explicit view pairs) and process the context view in a single forward pass with gradients that simultaneously yields diffusion logits and the JEPA embedding; only the target view requires an additional no-gradient forward pass. The result is a JEPA training step with 33\% fewer FLOPs than LLM-JEPA (Table~\ref{tab:compute}), standard bidirectional attention (no custom block-causal mask), and applicability to any text dataset.

    \item \textbf{A consistent empirical pattern: geometric--functional drift dissociation.} Layer-wise probing reveals that the DLLM-JEPA objective does not appear to minimize representation change---it redirects it. On GSM8K, DLLM-JEPA's hidden-state drift from the pre-trained weights is \emph{larger} than the baseline's (drift ratio 1.3--3.6$\times$ across configurations), yet its Wikitext functional forgetting is \emph{smaller}. The amplification concentrates in middle transformer layers---consistent with prior interpretations of middle layers as encoding compositional structure---and replicates on Dream-7B (ratio 1.28$\times$). We also empirically rule out representation collapse (\S\ref{sec:method}, Appendix~\ref{app:collapse}). We use ``dissociation'' in a descriptive sense throughout.

    \item \textbf{Joint task gain and base preservation.} On LLaDA-8B with the Wide-$t$ configuration ($t_L{=}0.1, t_H{=}0.9$, lr $1.4\times10^{-6}$, gentler than the aggressive $(0.2, 0.7)$ schedule used for the main task table), DLLM-JEPA simultaneously improves GSM8K accuracy \emph{and} drives Wikitext diffusion loss below the pre-trained base, while an L2-to-base parameter anchor achieves weak base preservation with zero task gain.

    \item \textbf{Consistent gains across 4 tasks $\times$ 2 backbones.} Under a single 4-shot evaluation protocol, DLLM-JEPA improves every (task, architecture) combination (Table~\ref{tab:main}). The 3-seed mean improvement is $+1.8$\,pp on LLaDA-8B GSM8K under Wide-$t$ ($\pm 0.4$ vs baseline $\pm 0.9$) and $+2.6$--$3.0$\,pp on aggressive-schedule tasks, where DLLM-JEPA tightens a $\pm 8.9$\,pp baseline spread to $\pm 3.9$\,pp. Best-seed numbers (up to $+18.7$\,pp) and available multi-seed statistics are in Appendix~\ref{app:multiseed}.
\end{itemize}

\paragraph{Scope of comparison.} We position LLM-JEPA as a structural motivator rather than a direct comparator: the two methods sit on different attention substrates, so we benchmark DLLM-JEPA against diffusion-only fine-tuning on the same backbones. Our focus is on how JEPA objectives can be naturally instantiated in masked diffusion LMs and how they affect fine-tuning dynamics.

\section{Related Work}

\paragraph{Masked Diffusion Language Models.}
Discrete diffusion models for text generation have gained attention as alternatives to autoregressive decoding~\citep{austin2021structured,lou2024discrete}.
LLaDA~\citep{nie2025llada} introduced a simple masked diffusion framework that achieves competitive performance with AR models at the 8B parameter scale.
MDLM~\citep{sahoo2024simple} and SEDD~\citep{shi2024simplified} provide further theoretical and empirical foundations.
These models employ bidirectional attention and learn to reverse a forward masking process, enabling parallel token generation at inference time.

\paragraph{Joint Embedding Predictive Architectures.}
I-JEPA~\citep{assran2023ijepa} demonstrated that predicting latent representations of masked image patches produces superior visual features compared to pixel-level reconstruction (MAE).
V-JEPA~\citep{bardes2024vjepa} extended this to video.
LLM-JEPA~\citep{huang2025llmjepa} adapted the framework to autoregressive language models, showing improvements across code generation and math reasoning tasks.
However, LLM-JEPA requires explicit view pairs and incurs significant computational overhead due to the causal attention constraint.

\paragraph{Fine-tuning Diffusion Language Models.}
The fine-tuning dynamics of masked diffusion language models remain relatively underexplored.
\citet{nie2025llada} demonstrate that supervised fine-tuning of LLaDA on instruction data yields strong instruction-following behavior, but do not analyze representation dynamics or the trade-off between task adaptation and base preservation.
Our work uses DLLM-JEPA as a lens to probe this trade-off and shows that representation-level regularization reshapes it qualitatively.

\section{Method}
\label{sec:method}

\subsection{Preliminaries: Masked Diffusion Language Models}

LLaDA~\citep{nie2025llada} defines a forward process that progressively masks tokens.
Given a clean sequence $x_0 = (x_0^1, \ldots, x_0^L)$, the forward process at time $t \in [0, 1]$ independently replaces each token with a special \texttt{[MASK]} token with probability $t$:
\begin{equation}
    q(x_t^i \mid x_0^i) = \begin{cases}
        x_0^i & \text{with probability } 1 - t \\
        \texttt{[MASK]} & \text{with probability } t
    \end{cases}
\end{equation}

The model $f_\theta$ learns to predict the original tokens from the masked sequence.
Letting $\mathcal{M}_t = \{i : x_t^i = \texttt{[MASK]}\}$ denote the set of masked positions at time $t$, the diffusion training objective is the per-token cross-entropy over masked positions,
\begin{equation}
    \Ldiff = \mathbb{E}_{t \sim \mathcal{U}(0,1),\, x_t \sim q(\cdot \mid x_0, t)} \left[ -\frac{1}{|\mathcal{M}_t|} \sum_{i \in \mathcal{M}_t} \log p_\theta(x_0^i \mid x_t) \right]
    \label{eq:diff}
\end{equation}

Crucially, $f_\theta$ employs \emph{bidirectional} attention: every token can attend to every other token, enabling rich contextual representations from a single forward pass.

\subsection{DLLM-JEPA: JEPA for Diffusion Language Models}

\begin{figure*}[t]
\centering
\includegraphics[width=\linewidth]{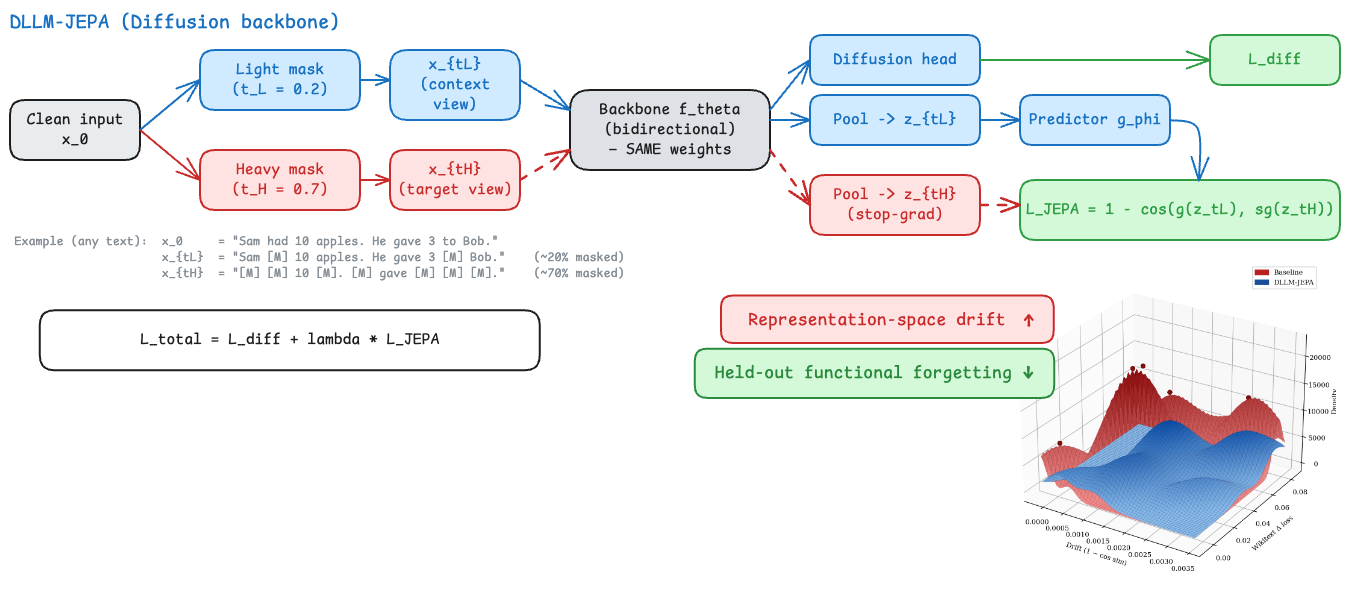}
\caption{\textbf{DLLM-JEPA at a glance.} \emph{Top row (training flow).} A single clean input $x_0$ is noised at two masking rates ($t_L{=}0.2$, $t_H{=}0.7$) to form a context view and a target view---no paired dataset required. The online backbone $f_\theta$ processes the context view in a single forward pass with gradients that yields diffusion logits (giving $\Ldiff$) and a pooled embedding $z_{t_L}$; the target view is processed by an EMA copy $f_{\theta'}$ (decay $\tau{=}0.996$) under \texttt{no\_grad} to produce $z_{t_H}$. A predictor $g_\phi$ maps $z_{t_L}{\to}\hat z_{t_H}$ and $\Ljepa$ is the cosine gap to $z_{t_H}$; $\Ltotal = \Ldiff + \lambda\Ljepa$ (cost 4F/step, Table~\ref{tab:compute}). A concrete masked-text example is shown in the middle-left. \emph{Bottom panels (observed mechanism).} Fine-tuning with DLLM-JEPA \emph{increases} hidden-state drift (1.36--3.60$\times$ baseline on GSM8K, concentrated in the middle transformer layers) yet \emph{reduces} Wikitext functional forgetting (43--58\%). The 3D density landscape visualizes the same effect over per-seed points: baseline mass peaks at high forgetting, while DLLM-JEPA mass extends along the drift axis yet stays closer to the zero-forgetting floor (\S\ref{sec:mechanism}).}
\label{fig:flow}
\end{figure*}

\paragraph{View generation via noise schedule.}
We generate two views of each input by sampling two masking rates from the diffusion schedule.
Given $t_L < t_H$ (we use fixed $t_L = 0.2$, $t_H = 0.7$), we construct:
\begin{align}
    \xtl &\sim q(x_t \mid x_0, t = t_L) \quad \text{(context view: 20\% masked)} \\
    \xth &\sim q(x_t \mid x_0, t = t_H) \quad \text{(target view: 70\% masked)}
\end{align}

The context view retains most tokens and provides a nearly complete picture, while the target view is heavily masked and forces prediction from a more incomplete, abstract context.
Both are generated from the \emph{same} input via different noise levels---no explicit view pairs needed.

\paragraph{Encoding and prediction.}
The backbone $f_\theta$ processes the context view $\xtl$ to produce hidden states, from which we obtain:
\begin{itemize}
    \item \textbf{Diffusion logits}: $\hat{x}_0 = \text{Classifier}(f_\theta(\xtl))$, used for the denoising objective.
    \item \textbf{JEPA embedding}: $\ztl = \text{Pool}(f_\theta(\xtl))$, a pooled representation of the context view. We use mean pooling over non-masked, non-padding tokens followed by LayerNorm; mean pooling is the standard sequence-to-vector reduction in vision JEPAs~\citep{assran2023ijepa,bardes2024vjepa}, and LayerNorm stabilizes the embedding scale---collapse prevention itself rests on the EMA target, stop-gradient, predictor, and the simultaneous diffusion denoising objective rather than on LayerNorm alone.
\end{itemize}

The target embedding $\zth = \text{Pool}(f_{\theta'}(\xth))$ is computed with an \emph{exponential moving average (EMA)} copy of the backbone, $f_{\theta'}$, following I-JEPA~\citep{assran2023ijepa} / BYOL~\citep{grill2020bootstrap} conventions. The EMA target is updated each step as $\theta' \leftarrow \tau\theta' + (1-\tau)\theta$ with decay $\tau{=}0.996$, and its forward pass runs under \texttt{no\_grad} so the target branch contributes no backward compute or optimizer state. The stop-gradient + EMA-target + predictor design provides a stable, slowly-moving prediction target and serves the standard role of preventing trivial-constant collapse, complementing SimSiam~\citep{chen2021simsiam}-style stop-gradient-only schemes.

A lightweight predictor $g_\phi$ consisting of $k$ transformer decoder layers (chosen for architectural consistency with LLM-JEPA's predictor design) maps the context embedding to predict the target:
\begin{equation}
    \hat{z}_{t_H} = g_\phi(\ztl)
\end{equation}

\paragraph{JEPA objective.}
Following best practices from vision JEPAs~\citep{assran2023ijepa}, we use cosine similarity:
\begin{equation}
    \Ljepa = 1 - \cos\!\big(\text{sg}(\zth),\; \hat{z}_{t_H}\big)
    \label{eq:jepa}
\end{equation}
where $\text{sg}(\cdot)$ denotes stop-gradient. Note that $\text{sg}(\zth)$ is redundant by construction (the target encoder is an EMA copy with \texttt{requires\_grad=False} and the forward is wrapped in \texttt{no\_grad}); we keep the notation for consistency with the BYOL/I-JEPA family of objectives.

\paragraph{Combined training objective.}
The total loss balances diffusion denoising and JEPA prediction:
\begin{equation}
    \Ltotal = \Ldiff + \lambda \cdot \Ljepa
    \label{eq:total}
\end{equation}
where $\lambda \geq 0$ controls the JEPA contribution and $k$ determines the predictor depth.

\paragraph{Empirical check: no representation collapse.}
Cosine-similarity-only JEPA objectives can in principle collapse to a trivial-constant solution. In our setting this is unlikely for two reasons: (i) the EMA target evolves slowly and provides a non-trivial prediction target; and (ii) the backbone is simultaneously optimized by the diffusion denoising loss $\Ldiff$, which constrains the token-level output distribution. We verify directly on the fine-tuned checkpoints (Appendix~\ref{app:collapse}): on LLaDA-8B ($D{=}4096$), the pooled JEPA embeddings $\ztl, \zth$ retain an effective rank of 42--44 (close to the pre-trained base's 42--43) with per-dimension standard deviation 0.73--0.95 and cosine diversity 0.25--0.28---virtually identical to the baseline fine-tuning on the same metrics. The pattern is the same on Dream-7B ($D{=}3584$, effective rank 43--44). The JEPA objective therefore does not reduce the rank or shrink the variance of the representation space; it \emph{redirects} the representation without collapsing it.

\subsection{Computational Advantage over LLM-JEPA}
\label{sec:compute}

\begin{table*}[t]
\centering
\caption{\textbf{Per-step training cost.} $F$: one forward pass through the backbone; $B$: one backward pass ($\approx 2F$ in FLOPs). LLM-JEPA requires gradient through both view encodings; DLLM-JEPA processes the target view without gradients.}
\label{tab:compute}
\small
\setlength{\tabcolsep}{6pt}
\begin{tabular}{@{}lccccc@{}}
\toprule
Method & Fwd (grad) & Fwd (no grad) & Backward & Total ($F$) & Overhead \\
\midrule
AR Baseline & 1$F$ & -- & 1$B$ $\approx$ 2$F$ & 3$F$ & -- \\
LLM-JEPA & 2$F$ & -- & 2$B$ $\approx$ 4$F$ & \textbf{6$F$} & +100\% \\
\midrule
Diffusion Baseline & 1$F$ & -- & 1$B$ $\approx$ 2$F$ & 3$F$ & -- \\
DLLM-JEPA (ours) & 1$F$ & 1$F$ & 1$B$ $\approx$ 2$F$ & \textbf{4$F$} & +33\% \\
\bottomrule
\end{tabular}
\end{table*}

Causal attention forces LLM-JEPA to encode the two views through \emph{two} forward passes with gradients under a custom block-causal mask. In contrast, the bidirectional attention in masked-diffusion backbones lets a single forward pass through $\xtl$ simultaneously yield the diffusion logits and the context embedding $\ztl$; the target embedding $\zth$ needs one additional forward pass through the EMA copy of the backbone, but this forward is gradient-free (no backward pass, no gradient memory, no optimizer state). The bookkeeping (Table~\ref{tab:compute}) gives $(6F{-}4F)/6F \approx \mathbf{33\%}$ fewer training FLOPs per step. The design also uses standard bidirectional attention, applies to any text dataset rather than paired (text, code) data, and adds only a pooling layer, a lightweight predictor, and one no-grad EMA forward to the vanilla diffusion training loop.

\section{Experiments}

We evaluate DLLM-JEPA on four tasks spanning code generation, SQL, regex synthesis, and math reasoning, using \emph{two} masked-diffusion backbones (LLaDA-8B and Dream-7B). Throughout, the comparison is DLLM-JEPA vs.\ diffusion-only fine-tuning on the same backbone; LLM-JEPA is used as the structural motivator rather than a direct head-to-head baseline.

\subsection{Experimental Setup}

\paragraph{Models.}
LLaDA-8B~\citep{nie2025llada} and Dream-7B~\citep{dream2025}, two recently released masked-diffusion language models with different pre-training recipes and tokenizers.

\paragraph{Tasks and metrics.}
\begin{itemize}
    \item \textbf{Django}~\citep{oda2015django}: 18,805 Python code generation pairs. Metric: whitespace-normalized prefix match (Appendix~\ref{app:dream-metrics}); 0-shot strict exact match also reported in Table~\ref{tab:django_decomp}.
    \item \textbf{Spider}~\citep{yu2018spider}: 7,000 text-to-SQL pairs. Metric: exact match and execution match on the Spider validation split (the official test set is hidden behind a leaderboard).
    \item \textbf{NL-RX-SYNTH}~\citep{locascio2016regex}: 10,000 text-to-regex pairs. Metric: exact match and functional equivalence (tested against 200 random strings).
    \item \textbf{GSM8K}~\citep{cobbe2021gsm8k}: 7,473 grade-school math problems with chain-of-thought answers. Metric: answer accuracy.
\end{itemize}

\paragraph{Training.}
All runs use full fine-tuning for 2 epochs on 8$\times$ NVIDIA A100 80GB GPUs with gradient checkpointing. We use AdamW with learning rate $1{\times}10^{-5}$ (primary) or $1.4{\times}10^{-6}$ (Wide-$t$ preservation-focused config), a fixed masking schedule $t_L{=}0.2, t_H{=}0.7$ for most experiments (Wide-$t$ uses $[0.1, 0.9]$), and DLLM-JEPA hyperparameters $\lambda \in \{0.5, 1.0, 2.0\}, k \in \{1,\ldots,5\}$.

\paragraph{Evaluation.}
We report 0-shot and 4-shot accuracy using each model's standard iterative unmasking generation with 128 diffusion steps. For Dream on code-structured tasks, whose natural output format may emit tokens after the answer, we additionally report the stopping-robust variant (whitespace-normalized prefix match for Django; truncation at the first \texttt{;} plus standard Spider cleaning for Spider) to enable fair cross-architecture comparison; raw metrics are reported in the appendix.

\subsection{Main Results: Task Performance}

Table~\ref{tab:main} presents DLLM-JEPA results across 4 tasks $\times$ 2 architectures.
Under the unified 4-shot protocol of Table~\ref{tab:main}, DLLM-JEPA improves every (task, architecture) cell, with gains ranging from $+1.00$ to $+18.73$ percentage points. A finer-grained 0-shot decomposition on LLaDA-8B Django (Table~\ref{tab:django_decomp}) shows a larger $+21.89$\,pp gain on the strict exact-match metric.

\medskip
\noindent\fcolorbox{gray!50}{gray!6}{\parbox{\dimexpr\linewidth-2\fboxsep-2\fboxrule}{%
\footnotesize\textbf{Reporting protocol (read once).} Table~\ref{tab:main} reports \emph{task-optimal} $(\lambda, k)$ per (task, architecture) cell under a unified 4-shot evaluation protocol using the \emph{aggressive} training schedule ($t_L{=}0.2, t_H{=}0.7$). A \emph{single-config} sibling (Table~\ref{tab:single_config}) repeats the comparison with $(\lambda{=}1, k{=}3)$ fixed across tasks; gains remain positive throughout but smaller. Multi-seed ($n{=}3$) statistics are in Table~\ref{tab:multiseed_main} (Appendix~\ref{app:multiseed}); LLaDA-8B Spider and Dream-7B Spider/NL-RX are single-seed grids over $(\lambda,k)$. Preservation results (\S\ref{sec:preservation}) use the \emph{Wide-$t$} schedule ($t_L{=}0.1, t_H{=}0.9$) chosen \emph{a priori}; base-capability probes are multi-seed on MMLU; Wikitext rows for Baseline and DLLM-JEPA are single-seed point estimates (the L2 anchor is reported over multiple seeds).}}
\medskip

\begin{table*}[t]
\centering
\caption{\textbf{Main results.} Task performance across four tasks and two masked-diffusion backbones, at task-optimal $(\lambda, k)$. Baselines and DLLM-JEPA share the same learning rate, training schedule, and seed within each row; $(\lambda, k)$ is not applicable to the baseline ($\lambda{=}0$). All rows use a single 4-shot protocol and the same metric for both models. Spider: SQL execution match with semicolon-truncation cleaning; Django: whitespace-normalized prefix match (Appendix~\ref{app:dream-metrics}). Fixed single $(\lambda, k)$ comparison in Table~\ref{tab:single_config}.}
\label{tab:main}
\small
\setlength{\tabcolsep}{5pt}
\begin{tabular}{@{}lllcccc@{}}
\toprule
Task & Metric (4-shot) & \multicolumn{2}{c}{LLaDA-8B} & \multicolumn{2}{c}{Dream-7B} \\
\cmidrule(lr){3-4} \cmidrule(lr){5-6}
 & & BL $\rightarrow$ JEPA & $\Delta$ & BL $\rightarrow$ JEPA & $\Delta$ \\
\midrule
GSM8K  & accuracy              & 42.61 $\rightarrow$ \textbf{61.33} & \textbf{+18.73} & 34.87 $\rightarrow$ \textbf{46.25} & \textbf{+11.38} \\
NL-RX  & func match            & 47.50 $\rightarrow$ \textbf{58.20} & \textbf{+10.70} & 42.00 $\rightarrow$ \textbf{46.80} & \textbf{+4.80}  \\
Spider & exec match (cleaned)  & 35.40 $\rightarrow$ \textbf{39.36} & \textbf{+3.97}  & 20.89 $\rightarrow$ \textbf{25.15} & \textbf{+4.26}  \\
Django & ws-prefix match       & 74.40 $\rightarrow$ \textbf{75.40} & \textbf{+1.00}  & 69.58 $\rightarrow$ \textbf{72.35} & \textbf{+2.77}  \\
\bottomrule
\end{tabular}
\end{table*}

Every (task, architecture) cell shows a positive improvement under a single 4-shot evaluation protocol and the same metric applied to both models. The LLaDA-8B GSM8K baseline of $42.6$ sits at the lower tail of a $\pm 8.9$ pp seed-dependent spread induced by the aggressive training schedule on diffusion-only fine-tuning; DLLM-JEPA collapses that spread to $\pm 3.9$ pp (Table~\ref{tab:multiseed_main}), so the cell-level gap reflects both peak improvement and variance reduction. Seeds are matched across methods within each row.

\paragraph{Cross-architecture replication (task performance).}
Dream-7B differs from LLaDA-8B in tokenizer, parameter count, and pre-training recipe. The same fine-tuning protocol applied to both yields the same direction of improvement on every task, supporting that the DLLM-JEPA effect is not an LLaDA-specific artifact; we emphasize that this is replication on one additional backbone, not a demonstration across all diffusion LMs.

\subsection{Base Capability Preservation}
\label{sec:preservation}

A central property of DLLM-JEPA is that task gain coexists with preservation of the base model on a held-out diffusion-loss probe. Preservation of pre-trained capability during task adaptation is the classic catastrophic-forgetting question~\citep{kirkpatrick2017overcoming}; parameter-space regularizers such as EWC are one response, while our analysis below offers a representation-space alternative. We evaluate this using held-out Wikitext-103 validation text, measuring the diffusion denoising loss ($\Ldiff$ with 50\% masking, deterministic seed) of each fine-tuned checkpoint against the pre-trained base. (DLLM-JEPA's Wikitext loss in fact ends up slightly \emph{below} the pre-trained base; we read this as evidence of preservation rather than a strong improvement claim, given that absolute Wikitext deltas are at the $\sim\!10^{-3}$ scale.)

\paragraph{Comparison with a parameter-anchor baseline.}
To show that DLLM-JEPA's benefit is not reducible to simple parameter proximity, we compare against an \textbf{L2-to-base anchor}: an SFT objective with an explicit penalty $\lambda_{L2}\lVert\theta - \theta_{\text{base}}\rVert_2^2$ (gradient-side injection for memory efficiency, $\lambda_{L2}{=}10^{-4}$). Both methods use the same Wide-$t$ configuration on LLaDA-8B GSM8K.

\begin{table*}[t]
\centering
\caption{\textbf{Task gain and base preservation together.} LLaDA-8B GSM8K, Wide-$t$ configuration. DLLM-JEPA is the only method that raises task accuracy while pushing Wikitext loss further below the base. The GSM8K accuracy column is over three seeds (per-seed breakdown in Table~\ref{tab:multiseed_gsm8k}); Wikitext rows for Baseline and DLLM-JEPA are single-seed (checkpoint-storage constraint), so we report a point estimate.}
\label{tab:preservation}
\small
\begin{tabular}{@{}lcc@{}}
\toprule
Method & GSM8K 0-shot & Wikitext $\Delta$ loss (vs.\ base) \\
\midrule
Base (no fine-tuning)                            & --     & 0.0000 \\
Diffusion Baseline ($\lambda{=}0$)               & 65.23 $\pm$ 0.93 & $-0.0004$ \\
L2-to-base anchor ($\lambda_{L2}{=}10^{-4}$)     & 65.18 $\pm$ 0.87 & $-0.0007 \pm 0.0002$ \\
\textbf{DLLM-JEPA} (ours)                        & \textbf{67.07 $\pm$ 0.41} & $\mathbf{-0.0017}$ \\
\bottomrule
\end{tabular}
\end{table*}

Three observations emerge. (1) The diffusion baseline achieves modest task gain and essentially matches the base in Wikitext loss. (2) The L2 anchor further reduces Wikitext drift slightly, but it does not improve task performance---its GSM8K score is statistically indistinguishable from the baseline. (3) DLLM-JEPA is the only method to achieve both: higher GSM8K accuracy (+1.84 pp over baseline) together with a Wikitext $\Delta$ below either competing method. The absolute Wikitext differences are small (${\sim}10^{-3}$), so readers should weigh them in combination with the task-accuracy column rather than in isolation.

\paragraph{MMLU sanity check (no catastrophic forgetting).}
As a sanity check that fine-tuning does not catastrophically degrade general capability, we also evaluate each Wide-$t$ checkpoint on \textbf{MMLU} (500 stratified questions; same items for every model), across three seeds (Table~\ref{tab:mmlu_preservation}). Both methods stay within about half a point of the base (baseline $57.93 \pm 0.42$, DLLM-JEPA $57.53 \pm 0.23$ vs.\ base $57.40$); the small mean gap is within seed noise, so we read MMLU as evidence \emph{against catastrophic forgetting on either method} rather than as a comparative claim. DLLM-JEPA's seed std is roughly half the baseline's; this matches the variance-tightening we observe on the high-variance LLaDA GSM8K cells but not on lower-variance cells. Our preservation argument primarily rests on the Wikitext evidence (Table~\ref{tab:preservation}).

\begin{table*}[t]
\centering
\caption{\textbf{MMLU sanity check.} 500-question stratified subset, three fine-tuning seeds. Both Wide-$t$ fine-tuning methods stay within about half a point of the base; the mean gap between Baseline and DLLM-JEPA is within seed noise.}
\label{tab:mmlu_preservation}
\small
\begin{tabular}{@{}lc@{}}
\toprule
Method & MMLU accuracy (\%) \\
\midrule
Base (no fine-tuning)          & 57.40 \\
Diffusion Baseline ($\lambda{=}0$) & 57.93 $\pm$ 0.42 \\
\textbf{DLLM-JEPA} (ours)       & \textbf{57.53 $\pm$ 0.23} \\
\bottomrule
\end{tabular}
\end{table*}

\section{Mechanism Analysis}
\label{sec:mechanism}

Having established that DLLM-JEPA delivers both task gains and base preservation, we next ask: \emph{what does the auxiliary objective do to the backbone internally?} We probe the fine-tuned checkpoints by comparing their layer-wise hidden-state representations against the pre-trained base on 100 Wikitext-103 samples. For each transformer layer we compute the mean-pooled representation and report the geometric drift $1-\cos\!\langle h^{\text{base}}_\ell, h^{\text{FT}}_\ell\rangle$ averaged over samples.

\begin{figure*}[t]
\centering
\includegraphics[width=\linewidth]{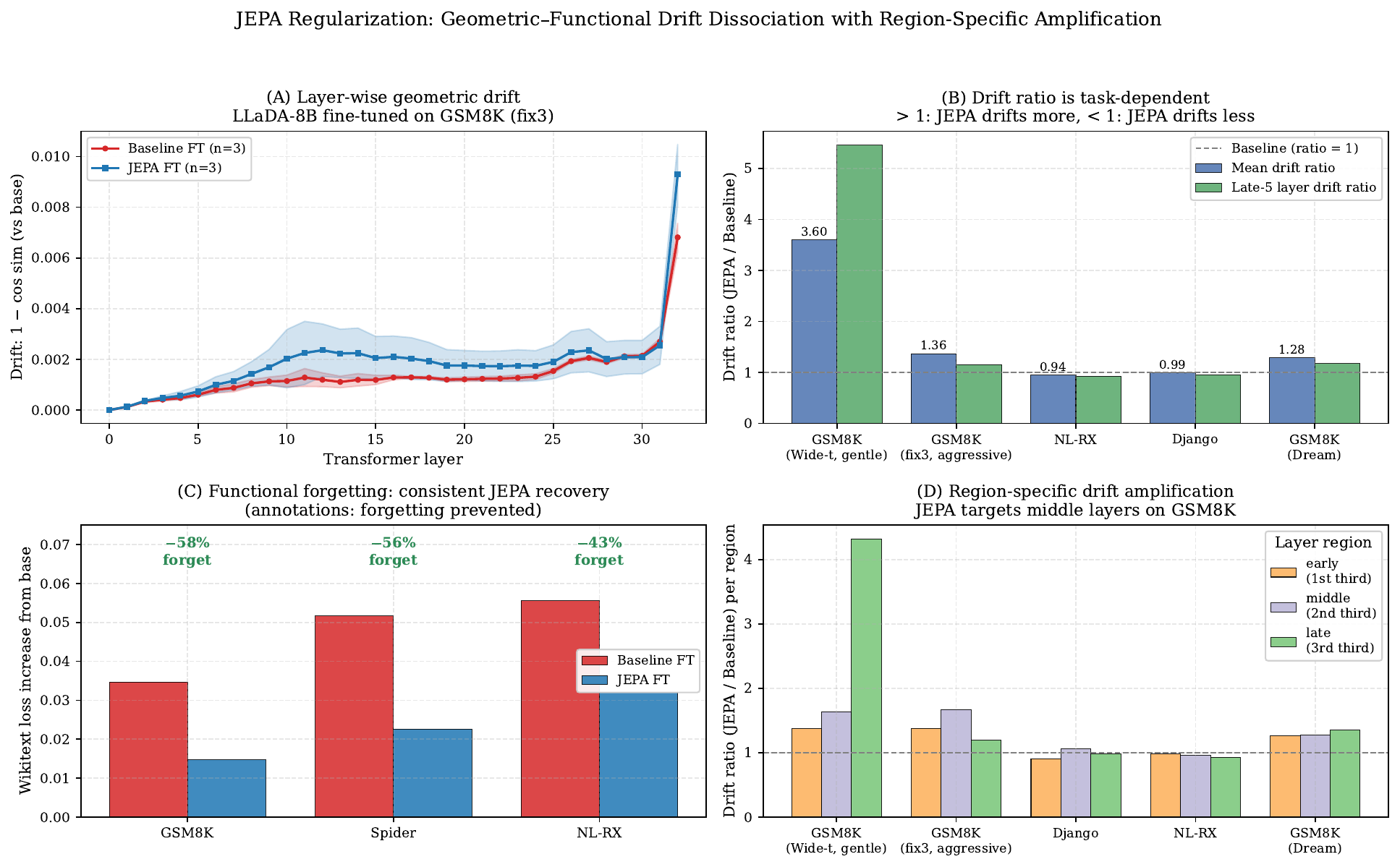}
\caption{\textbf{Geometric--functional drift dissociation for DLLM-JEPA.}
(A) Layer-wise geometric drift on LLaDA-8B fine-tuned on GSM8K (aggressive configuration). DLLM-JEPA drifts further from the base than the diffusion baseline throughout the network.
(B) Mean drift ratio (DLLM-JEPA / baseline) across five (task, configuration) cells. Amplification (ratio $>$ 1) is clear on the three GSM8K cells (both LLaDA configurations and Dream-7B); NL-RX and Django are near or slightly below 1, indicating task-specific amplification rather than uniform drift.
(C) Wikitext functional forgetting is \emph{reduced} for DLLM-JEPA on every fine-tuning task (43--58\%), opposite to the drift trend.
(D) Per-region drift ratio (early / middle / late thirds). On GSM8K, DLLM-JEPA's amplification concentrates in the middle layers, consistent with prior interpretations of middle-layer compositional structure.
All cells are multi-seed except LLaDA-8B GSM8K Wide-$t$ (single seed, checkpoint-storage constraint).}
\label{fig:hero}
\end{figure*}

\subsection{Geometric--Functional Drift Dissociation on GSM8K}
Figure~\ref{fig:hero}(A--B) reveals a counterintuitive finding: on GSM8K, DLLM-JEPA's hidden-state drift from the base is \emph{larger} than the baseline's---not smaller. On LLaDA-8B, the DLLM-JEPA / baseline drift ratio is $3.60\times$ with the gentle Wide-$t$ configuration and $1.36\times$ with the aggressive configuration. On NL-RX and Django (aggressive configuration) the ratio is $0.94\times$ and $0.99\times$ respectively---near 1, meaning DLLM-JEPA's drift amplification is \emph{task-specific}, not uniform. Yet Figure~\ref{fig:hero}(C) shows that on the three tasks where we measure Wikitext forgetting, DLLM-JEPA \emph{reduces} functional forgetting by 43--58\%.

This geometric--functional drift dissociation admits a descriptive interpretation: the JEPA objective does not appear to minimize representation change but instead redirects it along task-useful axes, leaving the base model's held-out output function largely intact.

\paragraph{Cross-architecture replication of the drift pattern.} The same direction of effect appears on Dream-7B (different parameter count, tokenizer, and pre-training corpus): DLLM-JEPA's drift ratio is $1.28\times$ (Figure~\ref{fig:hero}B, last column). Replication on one additional backbone is suggestive rather than conclusive.

\paragraph{Region-specific amplification (GSM8K).} The extra drift concentrates in the \emph{middle} transformer layers (Figure~\ref{fig:hero}D; ratios $1.64\times$ middle vs $1.38\times$ early / $1.18\times$ late under the aggressive schedule), which prior work associates with compositional reasoning. The pattern is sharp on GSM8K and muted on other tasks.

\subsection{Component ablation: asymmetric views and predictor}
\label{sec:decomposition}

\begin{figure*}[t]
\centering
\includegraphics[width=\linewidth]{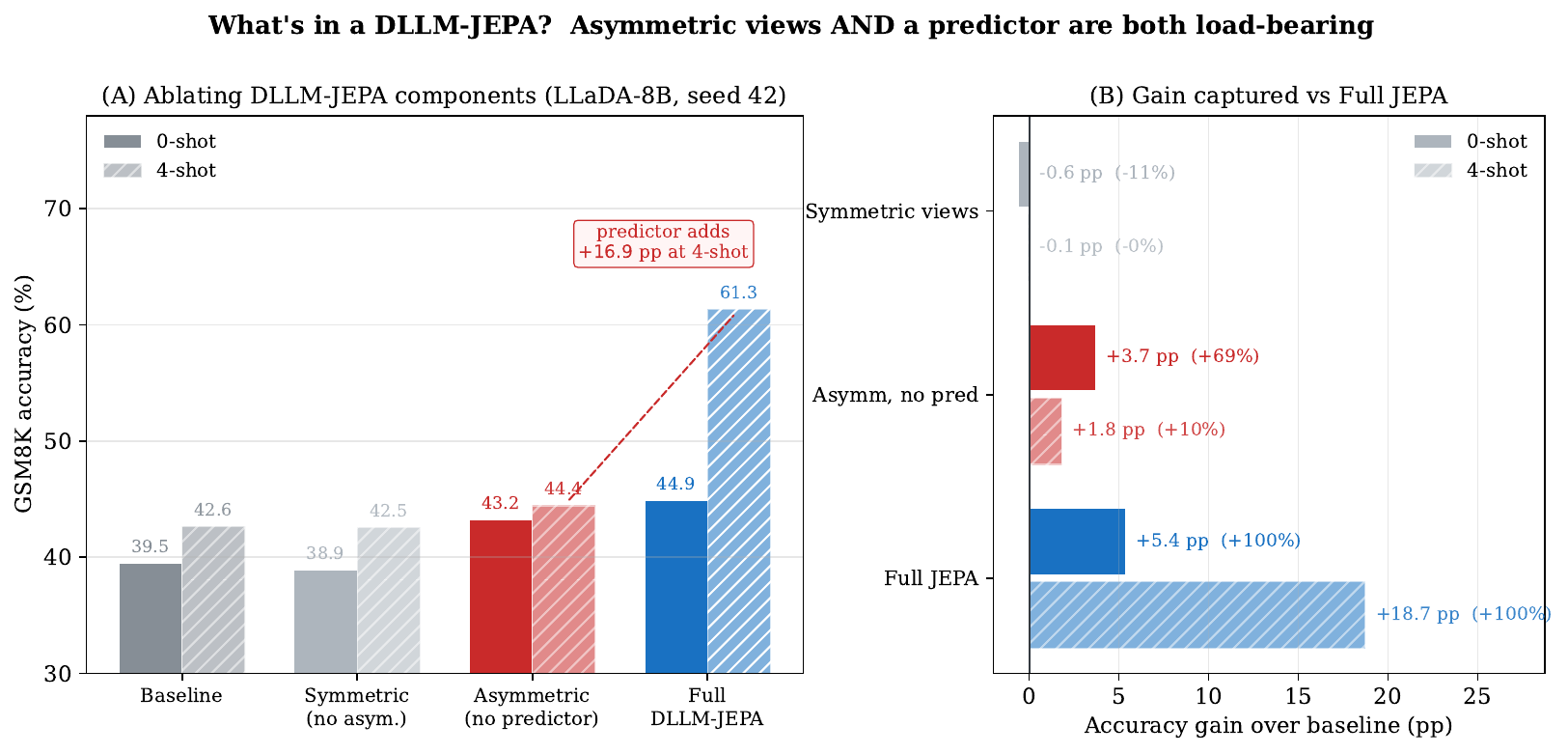}
\caption{\textbf{Ablating two DLLM-JEPA components.} (A) GSM8K accuracy on LLaDA-8B with aggressive fine-tuning. Removing asymmetric views collapses 0-shot to $38.9$\,pp --- \emph{below} the diffusion-only baseline --- and removing the predictor collapses 4-shot by $-16.9$\,pp relative to the full method. (B) Fractional gain over baseline: each partial configuration captures only a small (or negative) fraction of the full DLLM-JEPA gain.}
\label{fig:decomposition}
\end{figure*}

Figure~\ref{fig:decomposition} shows the effect of removing two DLLM-JEPA design choices on LLaDA-8B GSM8K (aggressive schedule, seed 42). (i) \textbf{Symmetric views} ($t_L{=}t_H{=}0.2$, same masking rate for both views) removes the asymmetric-noise structure of our method. (ii) \textbf{No-predictor} (identity map for $g_\phi$) removes JEPA's predictive head so the loss reduces to a plain stop-gradient two-view cosine consistency. The full DLLM-JEPA uses $(0.2, 0.7)$ and a $k$-layer predictor.

Two findings deserve emphasis. First, a symmetric-noise two-view setup, which would be a natural simplification, performs \emph{worse} than the diffusion-only baseline: the two views must probe different noise levels for the JEPA objective to produce a useful signal. Second, retaining the asymmetric views but discarding the predictor leaves much of the 0-shot gain intact but collapses the 4-shot gain, indicating that DLLM-JEPA is not reducible to a simple two-view consistency regularizer --- JEPA's \emph{predictive} structure contributes most of the few-shot benefit. Full noise-schedule sensitivity ($t_L, t_H$ grid on GSM8K) and additional ablations are in Appendix~\ref{app:ablations}.

Per-seed drift--forgetting scatter, joint density landscape, Django content/termination decomposition, and a structural comparison table to LLM-JEPA are in the appendix (\S\ref{app:scatter}--\S\ref{app:llmjepa-compare}).

\section{Limitations}

We identify three limitations that temper the strength of our conclusions:

\textbf{(L1) Scope of comparison.} We do not claim superiority over LLM-JEPA. The two methods sit on different attention substrates (causal vs.\ bidirectional), so a matched head-to-head would require aligning pre-training corpora, tokenizers, scales, and inference protocols---and even then the result would confound substrate with the JEPA objective itself. We therefore benchmark DLLM-JEPA against diffusion-only fine-tuning on the same backbone, and view the two methods as complementary instantiations of JEPA on different substrates.

\textbf{(L2) Preservation evidence.} Our preservation argument rests primarily on held-out Wikitext-103 diffusion loss (Table~\ref{tab:preservation}); the Wide-$t$ row is a single surviving seed (storage cleanup), so absolute deltas should be read alongside the direction-of-effect rather than as precise quantities. MMLU (Table~\ref{tab:mmlu_preservation}) is reported as a no-catastrophic-forgetting check, not a comparative claim---the small Baseline / DLLM-JEPA gap on MMLU is within seed noise.

\textbf{(L3) Statistical and operating-point scope.} The drift-dissociation and region-specific claims rest on three seeds per cell; the effect is largest on GSM8K, muted on NL-RX/Django, and replicated on one additional backbone (Dream). Our two configurations---aggressive $(t_L{=}0.2, t_H{=}0.7,$~lr~$10^{-5})$ and Wide-$t$ $(t_L{=}0.1, t_H{=}0.9,$~lr~$1.4\times10^{-6})$---are illustrative operating points, not the result of a fully factored $(\text{lr}, t_L, t_H)$ sweep, and all experiments are 2-epoch full-rank fine-tunings at the 7--8B scale; pre-training and other scales remain untested.

\section{Conclusion}

We introduced DLLM-JEPA, a JEPA formulation for masked-diffusion language models that cuts training FLOPs by 33\% relative to LLM-JEPA's architectural requirements while generating its two views automatically via the diffusion noise schedule. Under a single 4-shot protocol, DLLM-JEPA improves every (task, architecture) combination we evaluate, with gains up to +18.7\,pp on LLaDA-8B GSM8K and +11.4\,pp on Dream-7B GSM8K, and tightens run-to-run variance on the high-variance LLaDA-8B GSM8K aggressive-schedule baseline. Under the Wide-$t$ configuration ($t_L{=}0.1, t_H{=}0.9$, lr $1.4\times10^{-6}$) the method achieves task gain together with base preservation---raising GSM8K accuracy while driving Wikitext loss below the pre-trained base, unlike a parameter-level L2 anchor; MMLU stays close to base level on either method, with no evidence of catastrophic forgetting---and layer-wise probing uncovers a geometric--functional drift dissociation concentrated in middle transformer layers that replicates on Dream-7B. Future directions include applying DLLM-JEPA at pre-training, theoretical analysis of the dissociation mechanism, and adaptive scheduling of $\lambda$ during training; a matched-protocol comparison with LLM-JEPA is the natural next step.


\bibliographystyle{plainnat}

\begin{thebibliography}{99}

\bibitem[Assran et al.(2023)]{assran2023ijepa}
M.~Assran, Q.~Duval, I.~Misra, P.~Bojanowski, P.~Vincent, M.~Rabbat, Y.~LeCun, and N.~Ballas.
\newblock Self-supervised learning from images with a joint-embedding predictive architecture.
\newblock In \emph{CVPR}, 2023.

\bibitem[Austin et al.(2021)]{austin2021structured}
J.~Austin, D.~D. Johnson, J.~Ho, D.~Tarlow, and R.~van~den Berg.
\newblock Structured denoising diffusion models in discrete state-spaces.
\newblock In \emph{NeurIPS}, 2021.

\bibitem[Bardes et al.(2024)]{bardes2024vjepa}
A.~Bardes, Q.~Garrido, J.~Ponce, X.~Chen, M.~Rabbat, Y.~LeCun, M.~Assran, and N.~Ballas.
\newblock Revisiting feature prediction for learning visual representations from video.
\newblock \emph{arXiv preprint arXiv:2404.08471}, 2024.

\bibitem[Brown et al.(2020)]{brown2020language}
T.~Brown, B.~Mann, N.~Ryder, et~al.
\newblock Language models are few-shot learners.
\newblock In \emph{NeurIPS}, 2020.

\bibitem[Cobbe et al.(2021)]{cobbe2021gsm8k}
K.~Cobbe, V.~Kosaraju, M.~Bavarian, M.~Chen, H.~Jun, L.~Kaiser, M.~Plappert, J.~Tworek, J.~Hilton, R.~Nakano, C.~Hesse, and J.~Schulman.
\newblock Training verifiers to solve math word problems.
\newblock \emph{arXiv preprint arXiv:2110.14168}, 2021.

\bibitem[Devlin et al.(2019)]{devlin2019bert}
J.~Devlin, M.-W. Chang, K.~Lee, and K.~Toutanova.
\newblock {BERT}: Pre-training of deep bidirectional transformers for language understanding.
\newblock In \emph{NAACL}, 2019.

\bibitem[Grill et al.(2020)]{grill2020bootstrap}
J.-B. Grill, F.~Strub, F.~Altch\'e, C.~Tallec, P.~H. Richemond, E.~Buchatskaya, C.~Doersch, B.~{\'A}vila~Pires, Z.~D. Guo, M.~G. Azar, B.~Piot, K.~Kavukcuoglu, R.~Munos, and M.~Valko.
\newblock Bootstrap your own latent: A new approach to self-supervised learning.
\newblock In \emph{NeurIPS}, 2020.

\bibitem[Chen \& He(2021)]{chen2021simsiam}
X.~Chen and K.~He.
\newblock Exploring simple siamese representation learning.
\newblock In \emph{CVPR}, 2021.

\bibitem[Kirkpatrick et al.(2017)]{kirkpatrick2017overcoming}
J.~Kirkpatrick, R.~Pascanu, N.~Rabinowitz, J.~Veness, G.~Desjardins, A.~A. Rusu, K.~Milan, J.~Quan, T.~Ramalho, A.~Grabska-Barwinska, D.~Hassabis, C.~Clopath, D.~Kumaran, and R.~Hadsell.
\newblock Overcoming catastrophic forgetting in neural networks.
\newblock \emph{PNAS}, 114(13):3521--3526, 2017.

\bibitem[Huang et al.(2025)]{huang2025llmjepa}
H.~Huang, Y.~LeCun, and R.~Balestriero.
\newblock {LLM-JEPA}: Large language models meet joint embedding predictive architectures.
\newblock \emph{arXiv preprint arXiv:2509.14252}, 2025.

\bibitem[LeCun(2022)]{lecun2022path}
Y.~LeCun.
\newblock A path towards autonomous machine intelligence.
\newblock \emph{openreview.net preprint}, 2022.

\bibitem[Littwin et al.(2024)]{littwin2024jepa}
E.~Littwin, O.~Saremi, M.~Advani, V.~Thilak, P.~Nakkiran, C.~Huang, and J.~Susskind.
\newblock How {JEPA} avoids noisy features: The implicit bias of deep linear self distillation networks.
\newblock In \emph{NeurIPS}, 2024. arXiv:2407.03475.

\bibitem[Locascio et al.(2016)]{locascio2016regex}
N.~Locascio, K.~Narasimhan, E.~DeLeon, N.~Kushman, and R.~Barzilay.
\newblock Neural generation of regular expressions from natural language with minimal domain knowledge.
\newblock In \emph{EMNLP}, 2016.

\bibitem[Lou et al.(2024)]{lou2024discrete}
A.~Lou, C.~Meng, and S.~Ermon.
\newblock Discrete diffusion modeling by estimating the ratios of the data distribution.
\newblock In \emph{ICML}, 2024.

\bibitem[Nie et al.(2025)]{nie2025llada}
S.~Nie, F.~Zhu, Z.~You, X.~Zhang, J.~Ou, J.~Hu, J.~Zhou, Y.~Lin, J.-R. Wen, and C.~Li.
\newblock Large language diffusion models.
\newblock \emph{arXiv preprint arXiv:2502.09992}, 2025.

\bibitem[Sahoo et al.(2024)]{sahoo2024simple}
S.~S. Sahoo, M.~Arriola, Y.~Schiff, A.~Gokaslan, E.~Marroquin, J.~T. Chiu, A.~Rush, and V.~Kuleshov.
\newblock Simple and effective masked diffusion language models.
\newblock In \emph{NeurIPS}, 2024. arXiv:2406.07524.

\bibitem[Shi et al.(2024)]{shi2024simplified}
J.~Shi, K.~Han, Z.~Wang, A.~Doucet, and M.~Titsias.
\newblock Simplified and generalized masked diffusion for discrete data.
\newblock In \emph{NeurIPS}, 2024.

\bibitem[Touvron et al.(2023)]{touvron2023llama}
H.~Touvron, L.~Martin, K.~Stone, et~al.
\newblock Llama 2: Open foundation and fine-tuned chat models.
\newblock \emph{arXiv preprint arXiv:2307.09288}, 2023.

\bibitem[Yu et al.(2018)]{yu2018spider}
T.~Yu, R.~Zhang, K.~Yang, M.~Yasunaga, D.~Wang, Z.~Li, J.~Ma, I.~Li, Q.~Yao, S.~Roman, Z.~Zhang, and D.~Radev.
\newblock Spider: A large-scale human-labeled dataset for complex and cross-domain semantic parsing and text-to-{SQL} task.
\newblock In \emph{EMNLP}, 2018.

\bibitem[Ye et al.(2025)]{dream2025}
J.~Ye, Z.~Xie, L.~Zheng, J.~Gao, Z.~Wu, X.~Jiang, Z.~Li, and L.~Kong.
\newblock Dream 7{B}: Diffusion large language models.
\newblock \emph{arXiv preprint arXiv:2508.15487}, 2025.

\bibitem[Oda et al.(2015)]{oda2015django}
Y.~Oda, H.~Fudaba, G.~Neubig, H.~Hata, S.~Sakti, T.~Toda, and S.~Nakamura.
\newblock Learning to generate pseudo-code from source code using statistical machine translation.
\newblock In \emph{ASE}, 2015.

\bibitem[Merity et al.(2017)]{merity2017pointer}
S.~Merity, C.~Xiong, J.~Bradbury, and R.~Socher.
\newblock Pointer sentinel mixture models.
\newblock In \emph{ICLR}, 2017.

\bibitem[Hendrycks et al.(2021)]{hendrycks2021measuring}
D.~Hendrycks, C.~Burns, S.~Basart, A.~Zou, M.~Mazeika, D.~Song, and J.~Steinhardt.
\newblock Measuring massive multitask language understanding.
\newblock In \emph{ICLR}, 2021.

\end{thebibliography}

\newpage
\appendix
\section{Appendix}

\subsection{Per-Seed Drift--Forgetting Scatter and Density Landscape}
\label{app:scatter}

\begin{figure*}[t]
\centering
\includegraphics[width=0.6\linewidth]{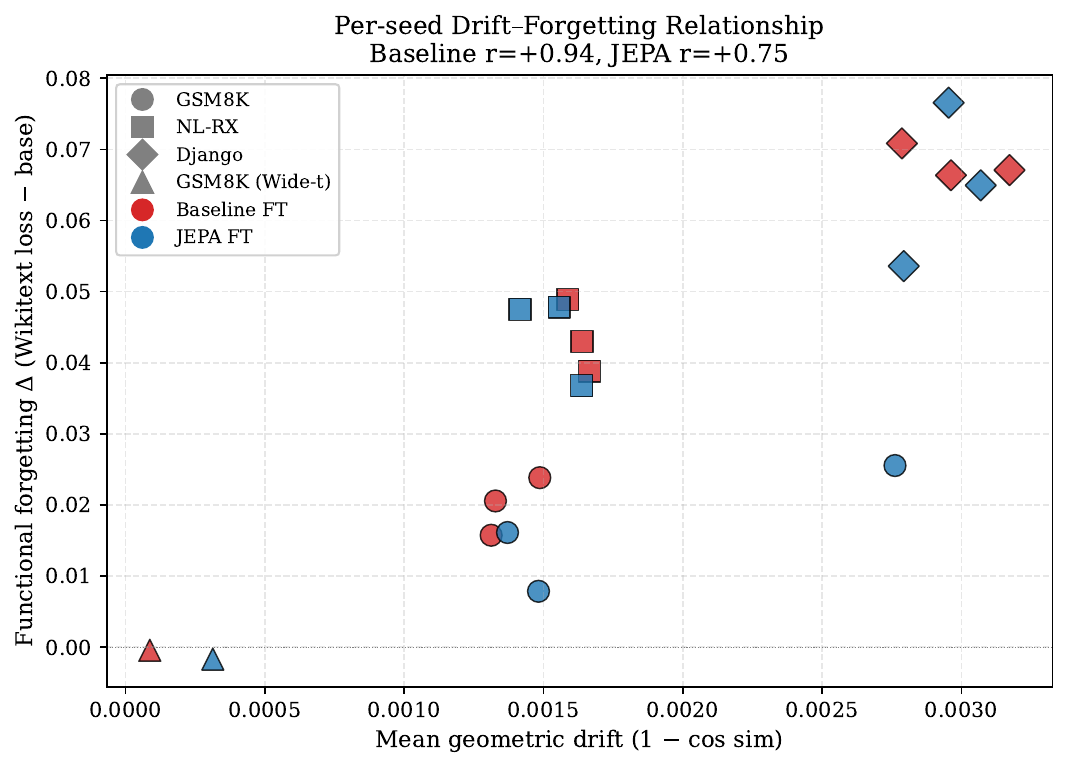}
\caption{\textbf{Per-run drift vs.\ functional forgetting.} One marker per fine-tuning run. Baselines show a strong positive relationship ($r{=}+0.94$); DLLM-JEPA visibly shifts below the baseline cluster at comparable drift levels, weakening the relationship ($r{=}+0.75$) and occupying region with lower forgetting at the same geometric drift.}
\label{fig:scatter}
\end{figure*}

Figure~\ref{fig:scatter} plots per-seed geometric drift against functional forgetting. Baseline runs show a strong positive drift--forgetting correlation ($r=+0.94$, 95\% CI $[0.76, 0.99]$, $n=10$); DLLM-JEPA runs sit below the baseline envelope at comparable drift levels, weakening the correlation to $r=+0.75$ (95\% CI $[0.23, 0.94]$, $n=10$). A Fisher $r$-to-$z$ test gives $Z=1.42$, $p=0.16$ at this sample size. We use ``dissociation'' for this weakened coupling rather than its elimination---DLLM-JEPA's $r$ is still positive with a CI excluding zero.

\begin{figure*}[t]
\centering
\includegraphics[width=0.5\linewidth]{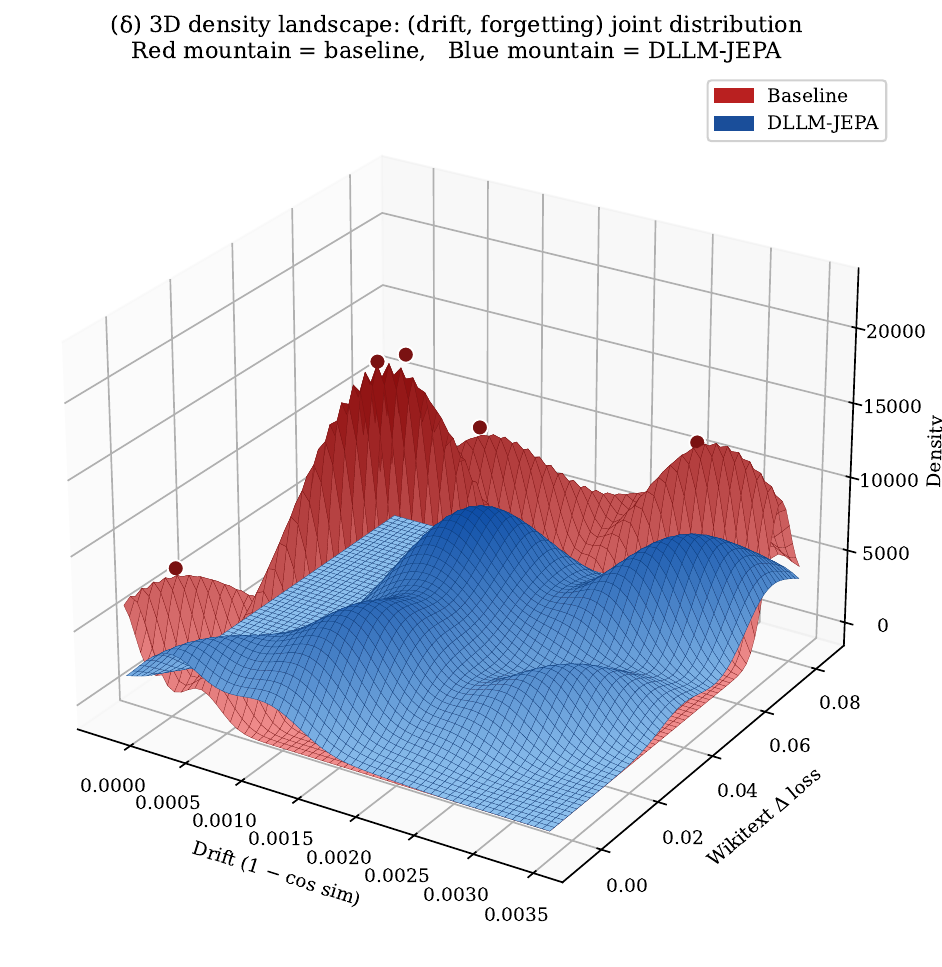}
\caption{\textbf{Joint (drift, forgetting) density landscape ($n{=}10$ runs per method, exploratory).} Gaussian-KDE surfaces over the same runs as Figure~\ref{fig:scatter}, baseline in red and DLLM-JEPA in blue. At matched drift, DLLM-JEPA's mass sits below the baseline's, and the baseline mass never reaches the low-forgetting corner that DLLM-JEPA occupies. Intended as an exploratory visualization, not a formal density estimate.}
\label{fig:drift_forget_density}
\end{figure*}

Figure~\ref{fig:drift_forget_density} re-plots the same data as a 3D density landscape. The baseline distribution is peaked at high forgetting (mass at Wikitext $\Delta\text{loss} \gtrsim 0.04$); DLLM-JEPA's mass stays closer to the $\Delta\text{loss}=0$ plane and extends further along the drift axis without a corresponding rise in forgetting.

\subsection{Content vs.\ Termination Decomposition on Django}
\label{app:django-decomp}

A finer-grained view of where the task gains come from is obtained by decomposing Django outputs into three categories: (i) exact match, (ii) ``stopping artifact'' (the correct code followed by trailing tokens), and (iii) genuine content error. For this decomposition we use the 0-shot exact-match setting on LLaDA-8B Django (single representative run).

\begin{table*}[t]
\centering
\caption{\textbf{Django 0-shot decomposition.} Single-seed illustrative run on LLaDA-8B. DLLM-JEPA simultaneously reduces stopping artifacts and reduces real content errors.}
\label{tab:django_decomp}
\small
\begin{tabular}{@{}lccc@{}}
\toprule
Category & Baseline & DLLM-JEPA & $\Delta$ \\
\midrule
Exact match                                    & 34.40\% & \textbf{56.29\%} & +21.89 \\
Stopping artifact (correct + trailing)         & 29.81\% & 11.30\%          & $-18.52$ \\
Content error                                  & 35.79\% & 32.41\%          & $-3.38$  \\
\midrule
Reasoning-correct total                        & 64.21\% & 67.59\%          & +3.38  \\
\bottomrule
\end{tabular}
\end{table*}

The exact-match improvement (+21.89pp) decomposes into an 18.52pp reduction in stopping artifacts and a 3.38pp improvement in underlying content quality. DLLM-JEPA's predictor objective provides an additional gradient signal about what comes next in representation space, translating into both sharper answer endings and cleaner content.

\subsection{Structural Comparison with LLM-JEPA}
\label{app:llmjepa-compare}

Table~\ref{tab:comparison} summarizes the architectural differences between LLM-JEPA~\citep{huang2025llmjepa} and DLLM-JEPA. The comparison is purely structural: these properties follow directly from the choice of substrate (causal vs.\ bidirectional attention), not from an empirical head-to-head. All task-performance claims in the main text are measured against diffusion-only fine-tuning.

\begin{table*}[t]
\centering
\caption{\textbf{Structural comparison: LLM-JEPA vs.\ DLLM-JEPA.} Architectural differences determined by the underlying substrate.}
\label{tab:comparison}
\small
\begin{tabular}{@{}lcc@{}}
\toprule
Property & LLM-JEPA & DLLM-JEPA (ours) \\
\midrule
Base architecture & Autoregressive & Masked Diffusion \\
Attention & Causal & Bidirectional \\
View construction & Explicit (Text $\leftrightarrow$ Code) & Automatic (noise schedule) \\
Requires view pairs & Yes & \textbf{No} \\
Forward passes (w/ grad) & 2 & \textbf{1} \\
Custom attention mask & Required & \textbf{Not needed} \\
Training FLOP overhead & +100\% & \textbf{+33\%} \\
Applicable datasets & Multi-view only & \textbf{Any text} \\
\bottomrule
\end{tabular}
\end{table*}

\subsection{Ablations}
\label{app:ablations}

\paragraph{$\lambda \times k$ grid.}
Table~\ref{tab:gsm8k_grid} shows GSM8K accuracy across the grid on LLaDA-8B; every non-zero $(\lambda, k)$ configuration outperforms the diffusion baseline. The best configuration is $(\lambda{=}2.0, k{=}4)$.

\begin{table*}[t]
\centering
\caption{\textbf{GSM8K $\lambda \times k$ grid.} 0-shot accuracy on LLaDA-8B. Baseline ($\lambda{=}0$): 0.402.}
\label{tab:gsm8k_grid}
\small
\begin{tabular}{@{}lccccc@{}}
\toprule
$\lambda$ \textbackslash\ $k$ & 1 & 2 & 3 & 4 & 5 \\
\midrule
0.5 & 0.445 & 0.450 & 0.457 & 0.459 & 0.438 \\
1.0 & 0.445 & 0.456 & 0.467 & 0.467 & 0.444 \\
2.0 & 0.437 & 0.460 & 0.473 & \textbf{0.488} & 0.453 \\
\bottomrule
\end{tabular}
\end{table*}

\paragraph{Single-configuration comparison.}
With a single fixed $(\lambda{=}1.0, k{=}3)$ applied across all tasks (Table~\ref{tab:single_config}), DLLM-JEPA still improves over the diffusion baseline on every row.

\begin{table*}[t]
\centering
\caption{\textbf{Single fixed configuration across tasks.}}
\label{tab:single_config}
\small
\begin{tabular}{@{}lllccc@{}}
\toprule
Task & Shot & Metric & Baseline & DLLM-JEPA & $\Delta$ \\
\midrule
Django & 0-shot & exact & 34.40 & 45.82 & +11.42 \\
NL-RX  & 4-shot & exact & 11.25 & 18.74 & +7.49 \\
Spider & 4-shot & exact & 25.73 & 29.21 & +3.48 \\
GSM8K (Wide-$t$) & 0-shot & acc & 65.28 & 66.87 & +1.59 \\
\bottomrule
\end{tabular}
\end{table*}

\paragraph{Noise schedule ablation.}
We use two illustrative training configurations:
(i) the \emph{aggressive} schedule with view-generation $(t_L{=}0.2, t_H{=}0.7)$ and lr $10^{-5}$, used for the main task table; and
(ii) the \emph{Wide-$t$} schedule with view-generation $(t_L{=}0.1, t_H{=}0.9)$ paired with a lower lr ($1.4\times10^{-6}$), used for the preservation/joint-improvement experiment in \S\ref{sec:preservation}.
The Wide-$t$ schedule increases DLLM-JEPA's drift ratio on GSM8K from $1.36\times$ to $3.60\times$ (Figure~\ref{fig:hero}B) while simultaneously achieving deeper base preservation (Table~\ref{tab:preservation}). These two schedules are illustrative operating points rather than the result of a fully factored $(\text{lr}, t_L, t_H)$ sweep; see Appendix~\ref{app:tlth-sens} for our partial sensitivity grid (within the aggressive lr).

\paragraph{$(t_L, t_H)$ sensitivity.}
\label{app:tlth-sens}
Table~\ref{tab:tlth_sens} sweeps the view-generation pair on LLaDA-8B GSM8K (aggressive schedule, seed 42). The $(0.2, 0.7)$ default dominates other pairs sharply on 4-shot (by $+14$ to $+19$\,pp) and more modestly on 0-shot. Wider ($0.3, 0.9$) and narrower ($0.1, 0.5$) pairs both degrade; the symmetric case $(0.2, 0.2)$ is reported in \S\ref{sec:decomposition} and drops below the diffusion-only baseline entirely. These results corroborate the component-decomposition finding that \emph{asymmetric} noise levels are load-bearing for DLLM-JEPA.

\begin{table}[t]
\centering
\caption{\textbf{$(t_L, t_H)$ sensitivity on LLaDA-8B GSM8K (aggressive, seed 42).}}
\label{tab:tlth_sens}
\small
\begin{tabular}{@{}lcc@{}}
\toprule
$(t_L, t_H)$                       & 0-shot & 4-shot \\
\midrule
$(0.1, 0.5)$                       & 43.59  & 44.05  \\
$(0.2, 0.5)$                       & 42.00  & 46.40  \\
$(0.2, 0.7)$ \textbf{(default)}    & \textbf{44.88}  & \textbf{61.33} \\
$(0.3, 0.9)$                       & 40.11  & 42.76  \\
\bottomrule
\end{tabular}
\end{table}

\paragraph{Multi-seed robustness.}
Table~\ref{tab:multiseed_main} reports 3-seed mean $\pm$ std. The direction matches Table~\ref{tab:main}, with smaller means due to seed variance in the diffusion-only baseline. On LLaDA-8B GSM8K (aggressive), the baseline's seed variance is $\pm 8.9$\,pp, which DLLM-JEPA narrows to $\pm 3.9$\,pp while remaining within the same upper envelope.

\begin{table*}[t]
\centering
\caption{\textbf{Multi-seed robustness.} Mean $\pm$ std over three fine-tuning seeds where available; std values are population standard deviations ($N$-denominator).}
\label{tab:multiseed_main}
\small
\setlength{\tabcolsep}{6pt}
\begin{tabular}{@{}lllc@{}}
\toprule
Model & Task & Baseline $\rightarrow$ DLLM-JEPA & $\Delta$ \\
\midrule
LLaDA-8B & GSM8K 4-shot (aggressive) & 55.14 $\pm$ 8.90 $\rightarrow$ 58.12 $\pm$ 3.91 & +2.98 \\
LLaDA-8B & GSM8K 0-shot (Wide-$t$)   & 65.23 $\pm$ 0.93 $\rightarrow$ 67.07 $\pm$ 0.41 & +1.84 \\
LLaDA-8B & NL-RX 4-shot (func)       & 45.93 $\pm$ 2.70 $\rightarrow$ 48.52 $\pm$ 4.55 & +2.58 \\
LLaDA-8B & Django 4-shot (ws-prefix) & 74.63 $\pm$ 0.16 $\rightarrow$ 74.83 $\pm$ 0.56 & +0.20 \\
Dream-7B & GSM8K 4-shot              & 37.65 $\pm$ 3.77 $\rightarrow$ 40.21 $\pm$ 4.34 & +2.56 \\
Dream-7B & Django 4-shot (ws-prefix) & 68.72 $\pm$ 1.72 $\rightarrow$ 69.81 $\pm$ 1.82 & +1.09 \\
\bottomrule
\end{tabular}
\end{table*}

\subsection{Full Grid Search Results}
\label{app:grids}

\begin{table*}[t]
\centering
\caption{\textbf{NL-RX-SYNTH $\lambda \times k$ grid.} 4-shot functional match on LLaDA-8B. Baseline: 0.475.}
\label{tab:nlrx_grid}
\small
\begin{tabular}{@{}lcccc@{}}
\toprule
$\lambda$ \textbackslash\ $k$ & 1 & 2 & 3 & 4 \\
\midrule
0.5 & 0.512 & 0.463 & 0.497 & 0.501 \\
1.0 & 0.465 & 0.432 & \textbf{0.582} & 0.534 \\
2.0 & 0.471 & 0.414 & 0.491 & 0.518 \\
\bottomrule
\end{tabular}
\end{table*}

Complete LLaDA-8B grids for Spider (both exact and execution match), NL-RX (exact, functional, and DFA-based functional match), and Django are released with the code.

\subsection{Multi-Seed Results}
\label{app:multiseed}

\begin{table*}[t]
\centering
\caption{\textbf{Multi-seed Wide-$t$ GSM8K.} LLaDA-8B 0-shot accuracy across three seeds.}
\label{tab:multiseed_gsm8k}
\small
\begin{tabular}{@{}lcccc@{}}
\toprule
Method & seed 42 & seed 123 & seed 777 & mean $\pm$ std \\
\midrule
Baseline FT             & 65.28 & 64.06 & 66.34 & 65.23 $\pm$ 0.93 \\
L2-to-base anchor       & 65.28 & 64.06 & 66.19 & 65.18 $\pm$ 0.87 \\
\textbf{DLLM-JEPA}      & \textbf{67.63} & \textbf{66.94} & \textbf{66.64} & \textbf{67.07 $\pm$ 0.41} \\
\bottomrule
\end{tabular}
\end{table*}

DLLM-JEPA's mean is higher than both baselines, with lower variance on this Wide-$t$ GSM8K cell specifically; aggregate variance behavior across all multi-seed cells is mixed (Table~\ref{tab:multiseed_main}). Multi-seed runs on the aggressive configuration across GSM8K, NL-RX, and Django are included in the released artifacts; the direction of improvement is consistent with the headline numbers reported in Table~\ref{tab:main}, though absolute values differ by task and configuration as expected.

\subsection{Representation-Collapse Diagnostic}
\label{app:collapse}

To empirically rule out representation collapse in our stop-gradient JEPA setting, we compute pooled context/target embeddings $\ztl = \mathrm{Pool}(f_\theta(\xtl))$ and $\zth = \mathrm{Pool}(f_\theta(\xth))$ on 100 Wikitext-103 samples for each checkpoint, then report: (i) mean per-dimension standard deviation, (ii) effective rank $r_{\mathrm{eff}} = (\sum_i \sigma_i)^2 / \sum_i \sigma_i^2$, and (iii) cosine diversity $1 - \overline{|\cos\langle z_i, z_j\rangle|}$ on 5{,}000 random pairs. Collapse would manifest as $r_{\mathrm{eff}} \to 1$ and cos-div $\to 0$.

\begin{table*}[t]
\centering
\caption{\textbf{Collapse diagnostic.} Measured on 100 Wikitext samples. For LLaDA-8B ($D{=}4096$) and Dream-7B ($D{=}3584$) across base / baseline-FT / DLLM-JEPA-FT checkpoints. Effective rank remains in the 41--44 range across all checkpoints, essentially matching the pre-trained base; cosine diversity and std are also preserved. No collapse is observed.}
\label{tab:collapse}
\small
\begin{tabular}{@{}llcccc@{}}
\toprule
Checkpoint & view & std & $r_{\mathrm{eff}}$ & cos-div \\
\midrule
LLaDA base                      & $\ztl$ & 0.76 & 42.3 & 0.28 \\
LLaDA base                      & $\zth$ & 0.95 & 41.8 & 0.27 \\
LLaDA baseline-FT (Wide-$t$)    & $\ztl$ & 0.77 & 42.8 & 0.28 \\
LLaDA baseline-FT (Wide-$t$)    & $\zth$ & 0.96 & 41.2 & 0.28 \\
\textbf{LLaDA DLLM-JEPA (Wide-$t$)} & $\ztl$ & 0.73 & 42.1 & 0.25 \\
\textbf{LLaDA DLLM-JEPA (Wide-$t$)} & $\zth$ & 0.95 & 41.8 & 0.27 \\
LLaDA baseline-FT (aggressive)        & $\ztl$ & 0.77 & 44.0 & 0.31 \\
\textbf{LLaDA DLLM-JEPA (aggressive)} & $\ztl$ & 0.75 & 42.4 & 0.28 \\
Dream base                      & $\ztl$ & 1.43 & 44.1 & 0.47 \\
Dream baseline-FT (GSM8K)       & $\ztl$ & 1.33 & 44.0 & 0.39 \\
\textbf{Dream DLLM-JEPA (GSM8K)} & $\ztl$ & 1.31 & 43.0 & 0.39 \\
\bottomrule
\end{tabular}
\end{table*}

\subsection{Reference Values from LLM-JEPA on Comparable Benchmarks}
\label{app:llmjepa-ref}

Table~\ref{tab:llmjepa-ref} lists the improvements reported in LLM-JEPA~\citep{huang2025llmjepa} alongside the improvements we obtain with DLLM-JEPA on the same task names. This is \textbf{not} a matched comparison---the two methods use different backbones (AR vs.\ masked-diffusion), different \emph{model scales} (1B vs.\ 7--8B), different pre-training corpora, different baseline recipes, and in one case a different metric variant. We present the table solely as a reference point for readers familiar with the LLM-JEPA paper; reading it as a direct DLLM-JEPA-vs-LLM-JEPA comparison would be incorrect. Dream-7B is not shown here because LLM-JEPA did not evaluate it.

\begin{table*}[t]
\centering
\caption{\textbf{LLM-JEPA reference numbers (not a matched comparison).} Values from LLM-JEPA~\citep{huang2025llmjepa} (Tables 2--4 therein) and DLLM-JEPA (this work), each measured against its own baseline. Note the model-scale gap (1B vs.\ 7--8B), different substrates, and the more lenient ``startswith'' metric used by LLM-JEPA on NL-RX-SYNTH.}
\label{tab:llmjepa-ref}
\footnotesize
\setlength{\tabcolsep}{4pt}
\begin{tabular}{@{}lllcc@{}}
\toprule
Task & Method (backbone, scale) & Metric & Baseline $\rightarrow$ method & $\Delta$ (pp) \\
\midrule
\multirow{3}{*}{GSM8K}
 & LLM-JEPA (Llama-3.2-1B-Instruct, 1B) & accuracy (3-seed)     & 32.4 $\rightarrow$ 36.4 & +4.0 \\
 & DLLM-JEPA (LLaDA-8B, Wide-$t$)       & 0-shot acc (3-seed)   & 65.2 $\rightarrow$ 67.1 & +1.8 \\
 & DLLM-JEPA (LLaDA-8B, aggressive)     & 4-shot acc (3-seed)   & 55.1 $\rightarrow$ 58.1 & +3.0 \\
\midrule
\multirow{2}{*}{Spider}
 & LLM-JEPA (Llama-3.2-1B-Instruct, 1B) & accuracy (3-seed)     & 47.5 $\rightarrow$ 50.6 & +3.0 \\
 & DLLM-JEPA (LLaDA-8B)                 & 4-shot exec (1 seed)  & 35.4 $\rightarrow$ 39.4 & +4.0 \\
\midrule
\multirow{2}{*}{NL-RX-SYNTH}
 & LLM-JEPA (Llama-3.1-8B-Instruct, 8B) & startswith\textsuperscript{\textdagger} & 35.8 $\rightarrow$ 63.6 & +27.8 \\
 & DLLM-JEPA (LLaDA-8B)                 & 4-shot func (3-seed)  & 45.9 $\rightarrow$ 48.5 & +2.6 \\
\bottomrule
\end{tabular}
\end{table*}

\textsuperscript{\textdagger}\emph{startswith} counts a prediction as correct if it begins with the gold regex string, irrespective of anything that follows; this is substantially more lenient than the functional-equivalence metric (our default), which tests the predicted regex against 200 randomly sampled strings and requires exact agreement with the gold regex's output. Neither baseline accuracy (35.8) nor final accuracy (63.6) under startswith is directly comparable to our functional-equivalence numbers. For orientation, on our LLaDA-8B checkpoints the startswith metric gives consistently higher absolute numbers than the functional metric, but this is not a substitute for evaluating LLM-JEPA under a matched protocol.

Setting the metric caveat aside, effect sizes on GSM8K and Spider fall in the same general range (single-digit percentage points over each paper's own baseline). Model-scale effects (LLM-JEPA at 1--8B, DLLM-JEPA at 7--8B) also factor into the absolute numbers; at a given scale, a larger model typically shows smaller absolute pp gains from the same auxiliary loss because its baseline is already closer to its ceiling.

\subsection{Stopping-Robust Evaluation Protocol}
\label{app:dream-metrics}

Dream-7B's natural generation for structured outputs (Python code, SQL) occasionally emits additional tokens after completing the answer, reflecting a generative style inherited from its pre-training. Under strict exact-match, this appears as a task failure despite the model having produced a correct answer. Because this is a model-style artifact rather than a quality difference, we adopt stopping-robust metrics for the tasks and models where this matters, and apply them \emph{uniformly} to both baseline and DLLM-JEPA, and to \emph{both} LLaDA-8B and Dream-7B in the main table (so column-to-column comparison is apples-to-apples):

\begin{itemize}
    \item \textbf{Django (ws-prefix):} whitespace-normalized prefix match between prediction and gold. On LLaDA-8B the ws-prefix and raw exact metrics are similar; on Dream-7B, ws-prefix avoids false negatives from trailing whitespace/docstrings.
    \item \textbf{Spider (exec-cleaned):} truncate prediction at the first semicolon, then run the standard Spider execution match against the database.
\end{itemize}

\paragraph{Raw vs.\ cleaned numbers for Dream.}
On Dream-7B Django 4-shot, raw exact match drops after DLLM-JEPA fine-tuning (e.g., baseline 12.1\% $\rightarrow$ JEPA 7.9\% on a representative run), which is entirely attributable to the stopping artifact: the model now more often emits a longer continuation including correct code plus additional tokens. Under ws-prefix, the same checkpoints show consistent positive gains (Table~\ref{tab:main}). On Dream-7B Spider 4-shot, raw exact/exec match are near zero for both baseline and JEPA (baseline 0.3\% / 0.6\%, JEPA 2.3\% / 5.2\%) due to the same mechanism; after SQL-cleaning, both rise into the 20--25\% range with DLLM-JEPA providing a clean +4.26 pp improvement.

\paragraph{0-shot results for completeness.}
The main-table 4-shot protocol is chosen for direct LLaDA--Dream comparability. 0-shot results include LLaDA-8B Django exact-match 34.4\% $\rightarrow$ 56.3\% (+21.9 pp), LLaDA-8B GSM8K Wide-$t$ 0-shot 65.3\% $\rightarrow$ 67.6\% (also reported in \S\ref{sec:preservation}), and Dream-7B NL-RX 0-shot functional match (DFA) 34.7\% $\rightarrow$ 42.4\% (+7.7 pp). These are not included in the main table only to preserve the unified 4-shot framing.

\subsection{Implementation and Reproducibility Details}
\label{app:repro}

\paragraph{Backbones.}
LLaDA-8B~\citep{nie2025llada}: 32 transformer layers, hidden dim 4096. Dream-7B~\citep{dream2025}: 28 layers, hidden dim 3584. Both use bidirectional attention. We use the public HuggingFace checkpoints and trust\_remote\_code.

\paragraph{JEPA module.}
The predictor $g_\phi$ is a stack of $k \in \{1,2,3,4,5\}$ transformer decoder layers with the same hidden dimension as the backbone, randomly initialized. Pooling is mean pooling over non-masked, non-padding tokens followed by a single LayerNorm. The EMA target encoder is a no-grad deepcopy of the backbone with decay $\tau{=}0.996$, updated every optimizer step.

\paragraph{Optimization.}
AdamW with $\beta_1{=}0.9, \beta_2{=}0.999$, weight decay $0.01$, cosine LR decay with 5\% warmup. Per-device batch size 2, gradient accumulation 2 (effective batch 32 across 8 GPUs). Gradient checkpointing enabled for memory. DeepSpeed ZeRO-2 (optimizer-state + gradient sharding across 8 GPUs).

\paragraph{Training schedules.}
\begin{itemize}
    \item \emph{Aggressive} (main task table): lr $1{\times}10^{-5}$, 2 epochs, fixed view rates $(t_L{=}0.2, t_H{=}0.7)$.
    \item \emph{Wide-$t$} (preservation): lr $1.4{\times}10^{-6}$, 2 epochs, fixed view rates $(t_L{=}0.1, t_H{=}0.9)$.
\end{itemize}
DLLM-JEPA hyperparameters: $\lambda \in \{0.5, 1.0, 2.0\}$ (loss weight on $\Ljepa$), $k \in \{1, 2, 3, 4, 5\}$ (predictor depth). Best $(\lambda, k)$ per cell reported in Table~\ref{tab:main}; full $k \in \{1,...,5\}$ grid for GSM8K in Table~\ref{tab:gsm8k_grid}, available $k \in \{1,...,4\}$ grid for NL-RX in Table~\ref{tab:nlrx_grid}.

\paragraph{Seeds.}
Multi-seed results use seeds $\{42, 123, 777\}$. Seeds are matched between baseline and DLLM-JEPA within each task / configuration cell so that any difference is attributable to the objective.

\paragraph{Hardware and wall-clock.}
$8 \times$ NVIDIA A100 80GB on a single node, 80\,GB HBM each. Per-run training wall-clock: GSM8K $\sim$45 min (LLaDA-8B aggressive), $\sim$1.0 h (Dream-7B aggressive), $\sim$50 min (LLaDA-8B Wide-$t$). Per-run evaluation wall-clock: GSM8K 0-shot $\sim$1.5 h, 4-shot $\sim$1.5 h (8-GPU DataParallel). Total compute for the experiments in this paper is approximately 800 A100-hours.

\paragraph{Generation / evaluation.}
LLaDA's iterative unmasking with 128 diffusion steps, block length 128, generation length 256 (GSM8K) or 512 (Django/NL-RX). Greedy decoding (temperature 0). Same generation config for baseline and DLLM-JEPA within each row.

\paragraph{Datasets.}
GSM8K~\citep{cobbe2021gsm8k}, Spider~\citep{yu2018spider}, NL-RX-SYNTH~\citep{locascio2016regex}, Django~\citep{oda2015django} from public sources. Wikitext-103 validation~\citep{merity2017pointer} for held-out diffusion loss. MMLU~\citep{hendrycks2021measuring} 500-question stratified subset (approximately 8--9 questions per subject across all 57 subjects, fixed seed) for general-capability check.

\paragraph{Code and artifacts.}
All training and evaluation scripts, configuration files, and the EMA + predictor + Wikitext probing code will be released at the paper's accompanying GitHub repository upon publication. The released bundle includes the exact $(\lambda, k)$ values per cell, masking-schedule scripts, and seed-by-seed checkpoints used for the multi-seed tables.

\end{document}